\documentclass{jpsj3}
\usepackage{txfonts}
\usepackage{comment}

\title{Information Compression and Performance Evaluation of Tic-Tac-Toe's Evaluation Function Using Singular Value Decomposition}

\author{Naoya Fujita and Hiroshi Watanabe\thanks{hwatanabe@appi.keio.ac.jp}}
\inst{Department of Applied Physics and Physico-Informatics, Keio University, Yokohama 223-8522, Japan} 

\abst{ We approximated the evaluation function for the game Tic-Tac-Toe by singular value decomposition (SVD) and investigated the effect of approximation accuracy on winning rate. We first prepared the perfect evaluation function of Tic-Tac-Toe and performed low-rank approximation by considering the evaluation function as a ninth-order tensor. We found that we can reduce the amount of information of the evaluation function by 70\% without significantly degrading the performance. Approximation accuracy and winning rate were strongly correlated but not perfectly proportional.
We also investigated how the decomposition method of the evaluation function affects the performance. We considered two decomposition methods: simple SVD regarding the evaluation function as a matrix and the Tucker decomposition by higher-order SVD (HOSVD). At the same compression ratio, the strategy with the approximated evaluation function obtained by HOSVD exhibited a significantly higher winning rate than that obtained by SVD. These results suggest that SVD can effectively compress board game strategies and an optimal compression method that depends on the game exists.}

\begin{document}
\maketitle

\section{Introduction}\label{sec:intro}

Game artificial intelligence (AI) is a computer program that plays board games, such as chess and Shogi and has been studied for a long time. In particular, computer chess has a long history. Computer chess programs that can even outperform humans have been developed~\cite{CAMPBELL200257}. However, since these programs were specialized for chess, they could not be generalized to other games. Recently, AlphaZero~\cite{Silver1140} has been gaining considerable attention as a general-purpose game AI. AlphaZero is a more generalized model of the AlphaGo Zero program~\cite{article111}, which demonstrated a higher performance than humans in Go by using a neural network (NN) to represent the rules of the game and being trained only through reinforcement learning from self-play. AlphaZero used a single network structure and defeated world champion programs in three different classical games, Go, chess, and Shogi, without any knowledge other than the rules of each game. Thus, general-purpose game AI can be created with high performance but the heuristic knowledge of the game. However, it is not possible to input all the information on the board directly into the NN for training, making it necessary to extract the features of the information on the board. In other words, some information compression is required. At present, the important information is extracted from the board heuristically, which is a crucial part of NN training. Therefore, a general method for compressing information on the board without any domain-specific knowledge is desired. One of the candidate information compression methods is singular value decomposition (SVD).

SVD is commonly used for information compression. It is a matrix decomposition method that allows low-rank approximation while retaining important information in the matrix. Therefore, it is often applied to reduce the number of parameters and compress the model size of NNs or tensor networks in fields such as image processing~\cite{6637309,inproceedings}, signal processing~\cite{8288893}, automatic speech recognition~\cite{article}, and quantum mechanics~\cite{8914525,PhysRevA.97.012327,PhysRevLett.115.180405}. However, this technique has not yet been applied to game AI, to our best knowledge.


In this study, we apply SVD to approximate the information on a game board and investigate the effect of approximation on a game AI's winning rate. We adopt Tic-Tac-Toe as the board game since the information space is small and we can search the entire game space. The board of Tic-Tac-Toe is a three-by-three grid. There are nine cells in total, and each cell takes on three different states. Thus, the state of the game board can be regarded as a ninth-order tensor. We first construct the perfect evaluation function for Tic-Tac-Toe and obtain approximated evaluation functions through low-rank approximation. Then, we investigate the relationship between the approximation accuracy and the game AI's winning rate. Since the evaluation function is a higher-order tensor, the decomposition is non-trivial. Thus, we consider two methods of decomposition, simple SVD and higher-order SVD (HOSVD)~\cite{ho,7070}. We compare the approximation accuracy and winning rate between the strategies approximated by simple SVD and HOSVD.

The rest of the article is organized as follows. The method is described in the next section. The results are shown in Sec.~\ref{sec:results}. Section~\ref{sec:summary} is devoted to summary and discussion.

\section{Method}\label{sec:method}

\subsection{Complete evaluation function}\label{subsec:method2}

\begin{figure}[htbp]
    \centering
    \includegraphics[width=3.5cm]{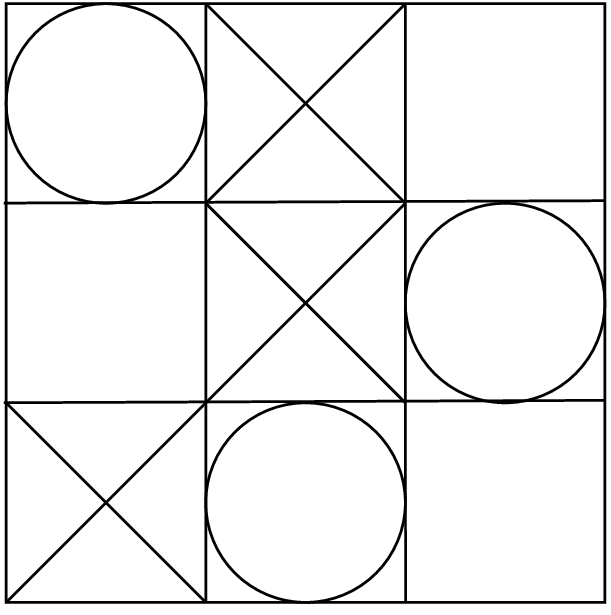}
    \caption{Typical state of Tic-Tac-Toe.}
    \label{fig:three_eyes1}
\end{figure}

Tic-Tac-Toe is a simple board game in which the first player plays with a circle and the second player plays with a cross on a $3 \times 3$ square board (Fig.~\ref{fig:three_eyes1})~\cite{708,298,800}. If a cell is not empty, it cannot be selected. The first player to place three of their objects in a row vertically, horizontally, or diagonally wins. The game is a draw if neither player can make a vertical, horizontal, or diagonal row. Tic-Tac-Toe is classified as a two-player, zero-sum, and perfect information game~\cite{RAGHAVA}.

In this paper, we refer to how much of an advantage either player has on a board as an evaluation value. The game AI examines the evaluation value from the information on the board and chooses the next move to increase the evaluation value. Therefore, it is necessary to define the board's evaluation value to construct the game AI's strategy. We refer to a function that returns an evaluation value of a given board as an evaluation function.

Suppose the current state of the board is $S$, which is the set of nine cell states. Each cell is numbered serially from $1$ to $9$. Then, a state is expressed as $S = \{c_1, c_2, \cdots, c_9\}$, where $c_i$ is the state of the $i$th cell and its value is $0$, $1$, and $2$ for empty, circle, and cross, respectively. The evaluation function $f(S)$ gives an evaluation value for a given state $S$. Since $S$ is the set of nine cell states and each cell can have three values, the evaluation function can be considered as a ninth-order tensor with dimension $3 \times 3 \times \cdots \times 3$.

Since the total number of states in Tic-Tac-Toe is at most $3^9 = 19~683$, even ignoring constraints and symmetries, we can count all possible states and construct the complete evaluation function. We refer to the evaluation function obtained by the full search as the perfect evaluation function $f_\mathrm{all}$. It is known that the game will always end in a draw if both players make their best moves. Thus, if we assume that both players choose the best move, the evaluated value of all states will be zero. Therefore, we calculate the complete evaluation value assuming that both players make moves entirely at random.

We first determine the evaluation values when the game is over. There are three terminal states in Tic-Tac-Toe: the first player wins, the second player wins, and the game is draw, with evaluation values of $1$, $-1$, and $0$, as follows,
\begin{equation}
    f_\mathrm{all}(S) = 
    \begin{cases}
        1  & \text{when the first player wins},  \\
        -1 & \text{when the second player wins}, \\
        0  & \text{when the game is draw}.       \\
    \end{cases}
\end{equation}

Next, we recursively define an evaluation value for a general state. We represent the state in which the game has progressed $n$ steps by $S_n$, \textit{i.e.}, $S_n$ contains $9-n$ empty cells. Suppose the $i$th cell is empty for a given state, \textit{i.e.}, $S_n = \{\cdots, c_i=0, \cdots\}$. Then we define the next state $S_{n+1}^i$ by replacing $c_i$ of $S_n$ as follows,
\begin{equation}
    S_{n+1}^i = 
    \begin{cases}
        \{\cdots, c_i=1, \cdots\}) & \text{when $n$ is even}, \\
        \{\cdots, c_i=2, \cdots\}) & \text{when $n$ is odd},
    \end{cases}
\end{equation}
since ($n+1$)th move is took by the first player when $n$ is even, while by the second player when $n$ is odd. Then the evaluation function for the state $S_n$ is given as the average for the all possible moves as
$$
    f_\mathrm{all}(S_n) = \frac{1}{9-n} \sum_i f_\mathrm{all}(S_{n+1}^i),
$$
where the summation is taken over all possible moves.

By repeating this process recursively from the initial state $S_0$, the state will reach one of the terminal states, the first player wins, the second player wins, and the game is draw. Then, the evaluation values for all the states are determined recursively.

\begin{figure}[ht]
    \centering
    \includegraphics[width=15cm]{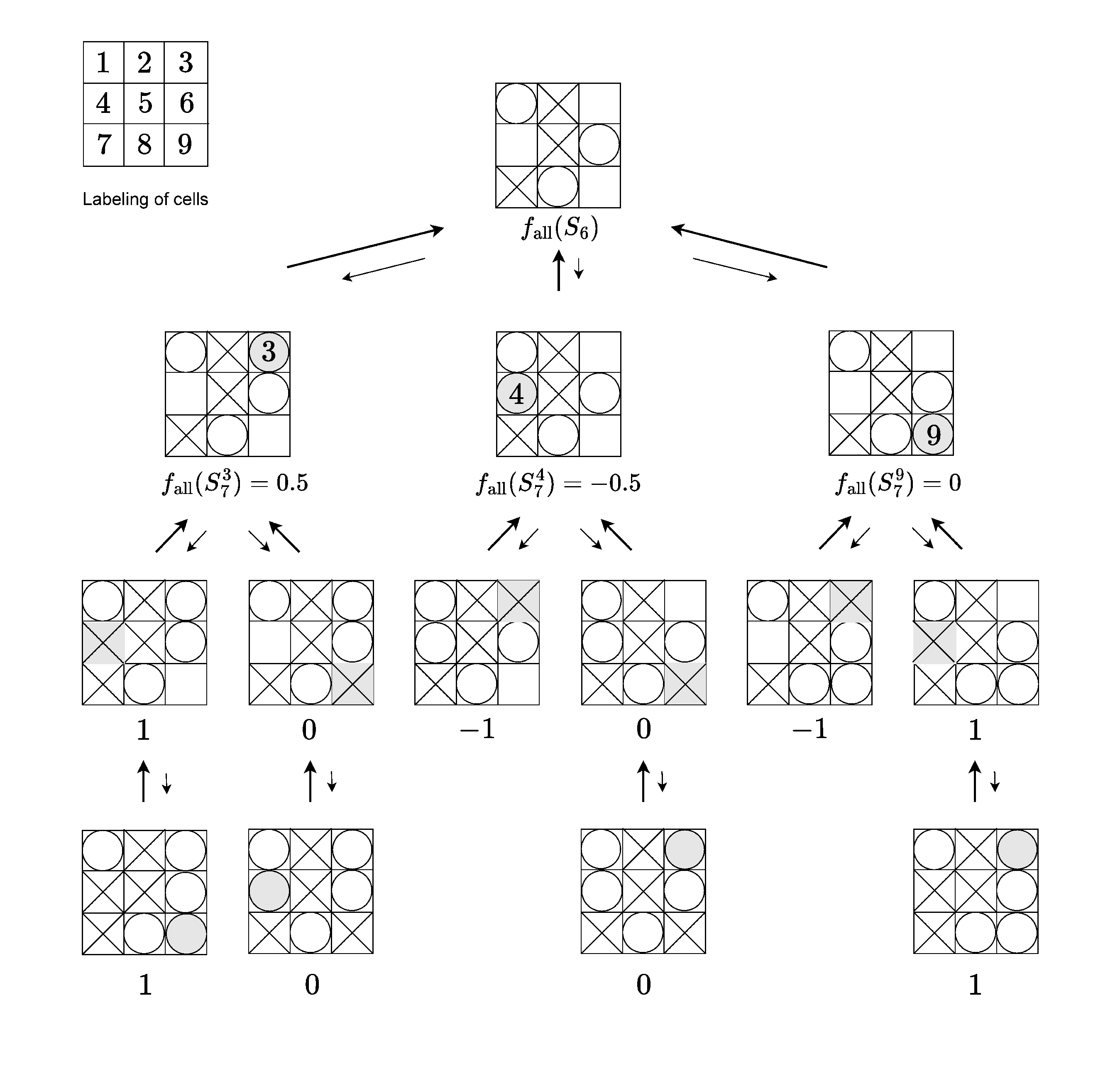}
    \caption{Calculation of the perfect evaluation function. The evaluation value of a given state is defined as the average of the evaluation values for the currently possible moves. The evaluation values are defined recursively. The evaluation value is defined as $1$ for a win, $-1$ for a loss, and $0$ for a draw.
    }
    \label{fig:three_eyes}
\end{figure}

An example of the recursive tree for determining the evaluation value of a state is shown in Fig.~\ref{fig:three_eyes}. The state $S_6$ contains six non-empty cells, and there are three possible next moves, cells 3, 4, and 9. The evaluation value $f_\mathrm{all}(S_6)$ of the current state $S_6$ is calculated as
\begin{equation}
    \begin{aligned}
        f_\mathrm{all}(S_6) & = \frac{1}{3} \left[ f_\mathrm{all}(S_7^3)+f_\mathrm{all}(S_7^4) + f_\mathrm{all}(S_7^9) \right] \\
                            & = \frac{1}{3}(0.5 - 0.5 + 0)                                                                     \\
                            & = 0 .
    \end{aligned}
\end{equation}
Here, all possible moves are equally weighted, which corresponds to the players choosing the next moves randomly. The closer the evaluation value is to $1$, the more likely the first player will win when both players choose a random move, and the closer it is to $-1$, the more likely the second player will win.

Note that, there are many invalid states in the tensor $f_\mathrm{all}$. Let $N_O$ be the number of marks for the first player and $N_X$ be the number of marks for the second player. Obviously, $N_O - N_X$ should be 0 or 1. But $f_\mathrm{all}$ contains the evaluation value for such invalid state which violates the condition $N_O - N_X = 0$ or $1$. Also, a state in which a mark is added after the game has already been completed is also invalid. For Tic-Tac-Toe, there are only $5~478$ valid states out of $3^9 = 19~683$ states. We set the evaluation values for such invalid states to be zero.

\subsection{Approximation of the evaluation function}

\begin{figure}[bthp]
    \centering
    \includegraphics[width=15cm]{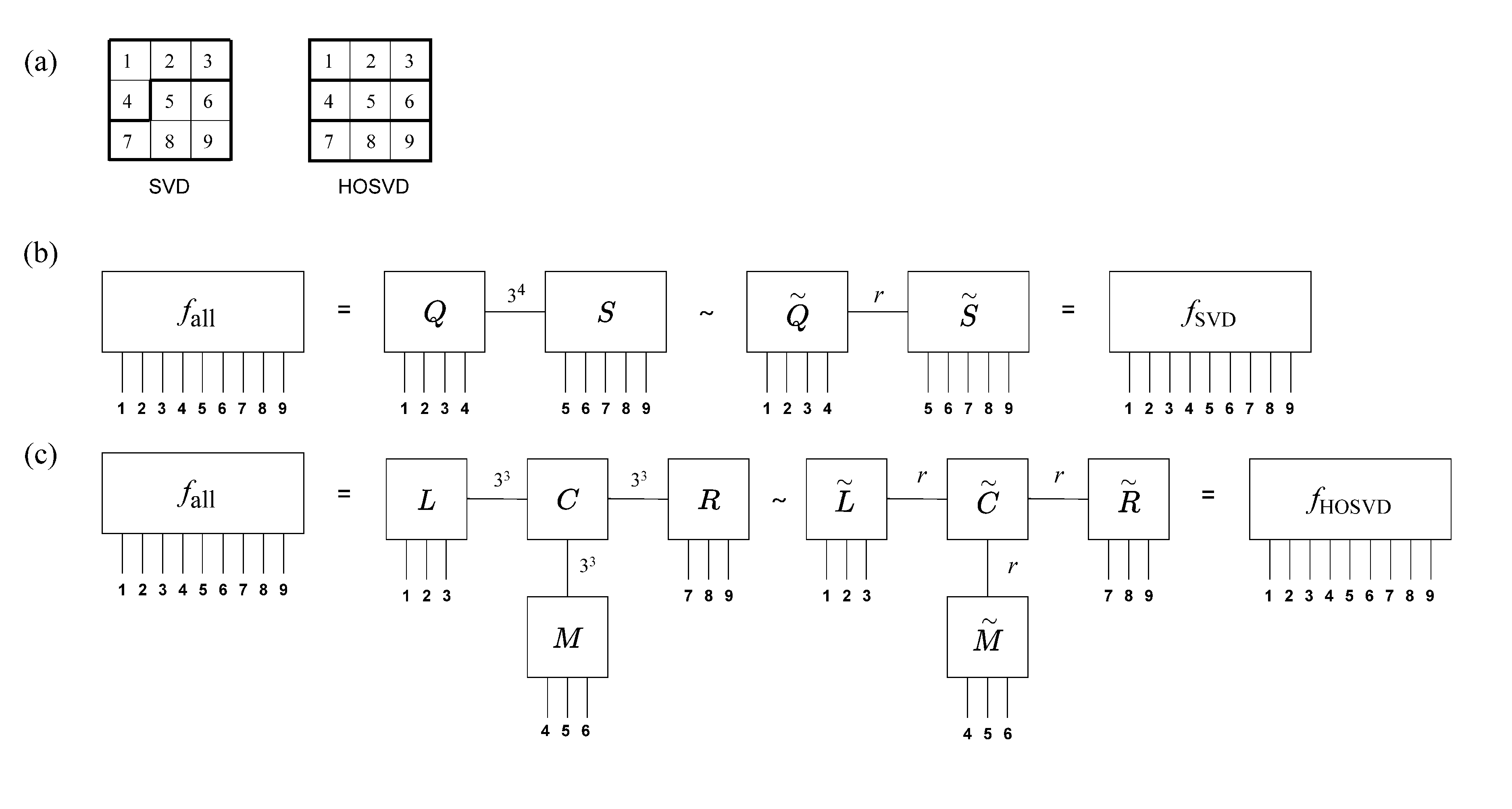}
    \caption{Decompositions and approximations of $f_{\mathrm{all}}$, which is a ninth-order tensor with dimension $3 \times 3 \times \cdots \times 3$. (a) Grouping of cell indices. The nine cells are divided into two groups, one containing cells $(1,2,3,4)$ and the other containing cells $(5,6,7,8,9)$ for the simple SVD. For HOSVD, they are divided into three groups, $(1,2,3)$, $(4,5,6)$, and $(7,8,9)$. (b) $f_{\mathrm{all}}$ is considered to be a matrix with dimension $3 ^ 4 \times 3 ^ 5$ and simple SVD is applied. (c) $f_{\mathrm{all}}$ is considered to be a third-order tensor with dimension $3 ^ 3 \times 3 ^ 3 \times 3 ^ 3$ and HOSVD is applied. Each index refers to the location of a cell. $r$ is the number of remaining singular values.}
    \label{fig:decomposition}
\end{figure}

The purpose of this study is to investigate how the approximation of the evaluation function $f_\mathrm{all}$ affects the winning rate. To approximate $f_\mathrm{all}$, we adopt SVD. However, the method of approximation is not uniquely determined since $f_\mathrm{all}$ is a higher-order tensor. We examined two approximation methods in the present study: simple SVD and HOSVD.

Since $f_\mathrm{all}$ is a ninth-order tensor with dimension $3 \times 3 \times \cdots \times 3$, it can be considered to be a $3^m \times 3^{9-m}$ matrix, where $m$ is an integer. Since the number of singular values, \textit{i.e.} the rank of the matrix, is determined by the smaller of $m$ and $9-m$, $m$ must be 4 or 5 to obtain as many singular values as possible. In this study, we set $m$ to $4$, \textit{i.e.}, we divide the nine cells into two groups, one containing four cells and the other containing five cells. The grouping of cells is shown in Fig.~\ref{fig:decomposition}~(a). We adopt the grouping $(1,2,3,4)$ and $(5,6,7,8,9)$ so that the length of the cut surfaces between groups became the shortest. The indices correspond to the position of cells which are denoted in Fig.~\ref{fig:three_eyes}.

Then the evaluation function $f_\mathrm{all}$ can be decomposed into two matrices $Q$ and $S$ by SVD as
\begin{eqnarray}
    f_{\mathrm{all}} &= U\Sigma V^* \equiv QS,\\
    Q & \equiv U\sqrt{\Sigma}, \\
    S & \equiv \sqrt{\Sigma}V^*,
\end{eqnarray}
where $U$ is a $3^4 \times 3^4$ matrix, $\Sigma$ is a $3^4 \times 3^5 $ rectangular diagonal matrix, $V^*$ is a $3^5 \times 3^5$ matrix, $Q$ is a $3^4 \times 3^4$ matrix, and $S$ is a $3^4 \times 3^5$ matrix, respectively. $V^*$ is the conjugate transpose of $V$. If we take $r$ singular values, $Q$ becomes a $3^4 \times r$ matrix $\tilde{Q}$ and $S$ becomes an $r \times 3^5$ matrix $\tilde{S}$. Then the approximated evaluation function is given by
\begin{equation}
    f_{\mathrm{all}} \sim f_{\mathrm{SVD}} = \tilde{Q} \tilde{S} .
\end{equation}
Schematic illustrations of this decomposition and approximation are shown in Fig.~\ref{fig:decomposition}~(b). 

As illustrated in Fig.~\ref{fig:decomposition}~(a), the grouping of cells for the simple SVD ignores the information on the game board. Since the purpose of Tic-Tac-Toe is to put three marks in a horizontal, vertical, or diagonal row, it is natural to adopt the grouping which reflects the rows of grid. Therefore, we divide the cells into three groups, which contains three rows or columns. Here, we divide the grid into three columns as shown in Fig.~\ref{fig:decomposition}~(a), \textit{i.e.}, the nine cells are divided into three groups, $(1,2,3)$, $(4,5,6)$, and $(7,8,9)$. Corresponding to this grouping, we considered the Tucker decomposition of the tensor. We regard $f_{\mathrm{all}}$ as a third-order tensor $X$ with dimension $3^3 \times 3^3 \times 3^3$. Then, the Tucker decomposition of the tensor $X$ is defined by
\begin{equation}
    X_{ijk} = \sum_{\alpha, \beta, \gamma} C_{\alpha \beta \gamma}
    L_{i \alpha}
    M_{j \beta}
    R_{k \gamma},
\end{equation}
where $L, M$, and $R$ are $3^3 \times 3^3$ matrices and $C$ is a third-order tensor with dimension $3^3 \times 3^3 \times 3^3$. The tensor $C$ is called a core tensor. See Fig.~\ref{fig:decomposition}~(c) for the graphical representation of this decomposition.

While the Tucker decomposition is not unique, we adopt HOSVD to determine the decomposition. First, we regard $f_{\mathrm{all}}$ as a matrix with dimension $3^6 \times 3^3$ with the grouping of the indices $(1,2,3,4,5,6)$ and $(7,8,9)$. Then the matrix $R$ is determined by the SVD of $f_{\mathrm{all}}$ as follows.
\begin{equation}
    f_{\mathrm{all}} =U_R \Sigma_R R,  \label{eq:R}
\end{equation}
where $U_R$ is a $3^6 \times 3^6$ unitary matrix, $\Sigma_R$ is a $3^6 \times 3^3$ rectangular diagonal matrix, and $R$ is a $3^3 \times 3^3$ unitary matrix which is what we want.

Next, we reorder the indices of $f_{\mathrm{all}}$ from $(1,2,3,4,5,6,7,8,9)$ to $(4,5,6,7,8,9,1,2,3)$. Then we again regard $f_{\mathrm{all}}$ as a matrix with dimension $3^6 \times 3^3$ with the grouping of the indices $(4,5,6,7,8,9)$ and $(1,2,3)$. From the SVD, we obtain the matrix $L$. We define the matrix $M$ with the similar procedure for the grouping $(1,2,3,7,8,9)$ and $(4,5,6)$.

From the obtained matrices $L$, $M$, and $R$, we define the core tensor $C$ as follows.
\begin{equation}
    C_{\alpha \beta \gamma} = \sum_{i, j, k} X_{i j k}
    L_{\alpha i }^*
    M_{\beta j}^*
    R_{\gamma k}^*.
\end{equation}

After the Tucker decomposition is obtained, we can obtain the approximated evaluation function by taking $r$ singular values. We define a matrix $\tilde{L}$ with the dimension $3^3 \times r$ by keeping only $r$ columns from the matrix $L$. Similarly, we define matrices $\tilde{M}$ and $\tilde{R}$. The approximated core tensor $\tilde{C}$ is obtained by
\begin{equation}
    \tilde{C}_{\alpha \beta \gamma} = \sum_{i, j, k} X_{i j k}
    \tilde{L}_{\alpha i}^*
    \tilde{M}_{\beta j}^*
    \tilde{R}_{\gamma k}^*,
\end{equation}
where $\tilde{C}$ is a third-order tensor with dimension $r \times r \times r$, and $\tilde{L}^*$ is a matrix with dimension $r \times 3^3$ which is the conjugate transpose of $L$. The same is true for $\tilde{M}^*$ and $\tilde{R}^*$. The approximated tensor $\tilde{X}$ is defined by
\begin{equation}
    \tilde{X}_{ijk} = \sum_{\alpha, \beta, \gamma}
    \tilde{C}_{\alpha \beta \gamma}
    \tilde{L}_{i \alpha}
    \tilde{M}_{j \beta}
    \tilde{R}_{k \gamma}.
\end{equation}
The $\tilde{X}$ is the third order tensor with the grouping of the indices $(1,2,3)$, $(4,5,6)$, and $(7,8,9)$. Reshaping the tensor $X$ to the ninth order tensor, we obtain the approximated evaluation function $f_{\mathrm{HOSVD}}$. Schematic illustrations of this decomposition and approximation are shown in Fig.~\ref{fig:decomposition}~(b).

\subsection{Compression ratio and relative error}

We introduce the compression ratio $Cr$ and the relative error $E$ to evaluate the quality of the approximations. $Cr$ is the ratio of the total number of elements in the approximated tensor to the number of elements in the original tensor. Suppose matrix $X$ is approximated as $X \simeq \tilde{Q}\tilde{S}$, then the compression ratio is defined as
\begin{equation}
    Cr = \frac{N(\tilde{Q}) + N(\tilde{S})}{N(X)} ,
\end{equation}
where $N(X)$ is the number of elements in matrix $X$. We define the compression ratio as high when $Cr$ is small and low when $Cr$ is high. For simple SVD, the tensor with $3^9$ elements is approximated by two matrices with dimensions $3^4 \times r$ and $r \times 3^5$. Therefore, the $r$ dependence of the compression ratio is
\begin{equation}
    Cr(r) = \frac{(3^4+3^5)r}{3^9} = \frac{4r}{3^5}.    
\end{equation}
Since $r$ ranges from $0$ to $81$, the compression ratio of the non-approximated evaluation function is $Cr=4/3$, which is greater than $1$. Note that, the matrix $X$ contains many zero elements which correspond to the irrelevant states to the game. On the other hand, matrices $\tilde{Q}$ and $\tilde{S}$ generally do not contain zero elements. Therefore, SVD degrades the compressibility since the redundancy of the original matrix will be lost.

We can define the compression ratio for HOSVD in a similar way. Suppose the tensor $X$ is approximated as $X \simeq \tilde{C} \times \tilde{L} \times \tilde{M} \times \tilde{R}$, where $\tilde{C}$ is the core tensor with the dimension $r \times r \times r$ and $\tilde{L}$, $\tilde{M}$, and $\tilde{R}$ are the matrices with the dimension $3^3 \times r$. Then the total number of elements are $3^4r + r^3$. Therefore, the $r$ dependence of the compression ratio for HOSVD is
\begin{equation}
    Cr(r) = \frac{N(\tilde{C})+N(\tilde{L})+N(\tilde{M})+N(\tilde{R})}{N(X)} = \frac{3^4 r + r^3}{3^9}.
\end{equation}

We define the relative error $E$ using the Frobenius norm. Suppose $X$ is the original tensor and $\tilde{X}$ is an approximated tensor. Then the relative error $E$ is define as
\begin{equation}
    E = \frac{\parallel X - \tilde{X} \parallel}{\parallel X \parallel} ,
\end{equation}
where $\parallel X \parallel$ is the Frobenius norm of the tensor $X$. With this definition, the compression ratio dependence of the relative error is equivalent to the singular value distribution of the original tensor.

\subsection{Strategy of game AI}

The game AI stochastically chooses the next move on the basis of the complete or approximated evaluation function. Suppose the current state of the board is $S_n$ and the evaluation function of the game AI is $f$. The evaluation value when the position of the next move is the $i$th cell is denoted by $f(S_{n+1}^i)$. The AI will choose among the possible moves that can be played next, according to their weight as follows. For visibility, we denote $f(S_{n+1}^i)$ as $\alpha_i$. Then the probability of choosing the $i$th cell for the next move, $p_i$, is determined by a softmax-type function~\cite{NIPS,max} as
\begin{equation}
    p_i = \frac{\mathrm{exp}(w \alpha_i)}{\sum_{j}\mathrm{exp}(w \alpha_j)},
\end{equation}
where $w$ is a parameter that determines how much the weight is emphasized and the summation is taken over all possible moves. The game AI chooses the cell with the largest evaluation value more frequently as $w$ increases. When $w$ is $0$, the evaluation value is ignored, and the game AI chooses the next move randomly. Therefore, the parameter $w$ plays the role of the inverse temperature of a Boltzmann weight. We choose $w = 10$ throughout the present study. We also simulated with other temperature values and confirmed that the qualitative behavior did not change. Since the evaluation value is set to $1$ when the first player wins and $-1$ when the second player wins, we adopt $-f(S)$ for the evaluation function for the second player.

While the evaluation values for the invalid states are zero in the original evaluation function $f_\mathrm{all}$, they are deviated from zero for the approximated evaluation functions $f_\mathrm{SVD}$ and $f_\mathrm{HOSVD}$. However, the AI selects the next move only from valid states, so the evaluation values of invalid states are never referenced.

\section{Results}\label{sec:results}

We allow game AIs to play games with each other with evaluation functions compressed at various compression ratios. We perform $500$ games in each case, switching the first and second players. Each player assumes that the opponent adopts the same evaluation function. The compression ratio and the number of remaining singular values are summarized in Table~\ref{table:compression}.

\subsubsection{Rank dependence of winning rates}

\begin{figure}[htbp]
    \centering
    \includegraphics[width=15cm]{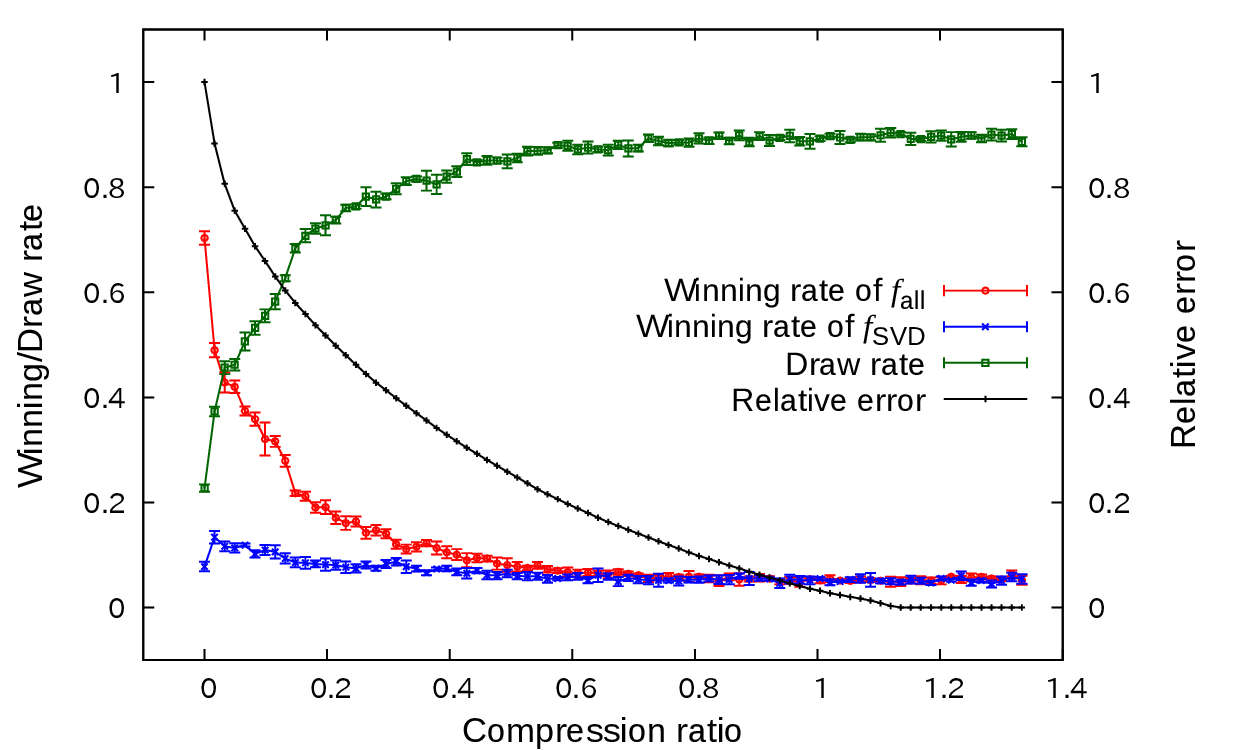}
    \caption{(Color online) Winning rates of game AIs with the evaluation functions $f_{\mathrm{all}}$ (red) and $f_{\mathrm{SVD}}$ (blue). The winning rates are almost constant down to a compression ratio of $0.3$. The relative error (black) is also shown. The winning rate is not perfectly proportional to the relative error.}
    \label{fig:res1}
\end{figure}

We first compare the complete evaluation function $f_\mathrm{all}$ and the evaluation function approximated by simple SVD, $f_\mathrm{SVD}$, to investigate the effect of low-rank approximation on the winning rate. Since we consider $f_\mathrm{all}$ as a matrix with dimension $3^4 \times 3^5$, the maximum number of singular values is $81$. Therefore, we examine the winning rate by varying the rank from $0$ to $81$. The winning and draw rates as functions of the compression ratio are shown in Fig.~\ref{fig:res1}.

One can see that the winning rates of $f_\mathrm{all}$ and $f_\mathrm{SVD}$ are almost constant down to a compression ratio of $0.3$. This compression ratio is achieved when $r=18$. The original evaluation function $f_\mathrm{all}$ contains $3^9$ numbers. If we adopt the double precision number to express the numbers, it will require 154 KB to store in memory. The approximated evaluation function $f_\mathrm{SVD}$ can be stored as two matrices, $\tilde{Q}$ and $\tilde{S}$. Since $\tilde{Q}$ is the $3^4 \times r$ matrix and $\tilde{S}$ is the $r\times 3^5$ matrix, two matrices contain 5~832 numbers in total when $r=18$, which requires 46 KB. Since the original evaluation function occupies 153 KB in memory while the approximated evaluation function occupies 46KB, we can reduce the amount of data of the evaluation function by 70\% without performance degradation. The relative error is also shown in Fig.~\ref{fig:res1}. As the compression ratio decreases, the relative error increases as expected. However, it is not entirely proportional to the winning rate of the game AI with the approximated evaluation function. Although the winning rate of $f_{\mathrm{all}}$ increases sharply when the compression ratio is lower than $0.3$, the relative error changes gradually. Since the relative error is the sum of the ignored singular values, this result shows that the performance of the evaluation function of a board game does not entirely depend on the singular value distribution.

\subsubsection{Dependence of decomposition methods}

\begin{table}[htbp]
    \centering
    \caption{Compression ratio and number of remaining singular values}
    \begin{tabular}{|c||c|c|c|c|c|c|c|c|} \hline
        Compression ratio                                 & 0.0 & 0.049 & 0.13 & 0.20 & 0.31 & 0.43 & 0.80 & 1.0 \\ \hline
        Number of singular values of $f_{\mathrm{SVD}}$   & 0   & 3     & 8    & 12   & 19   & 26   & 49   & 61  \\ \hline
        Number of singular values of $f_{\mathrm{HOSVD}}$ & 0   & 7     & 12   & 14   & 17   & 19   & 24   & 26  \\ \hline
    \end{tabular}
    \label{table:compression}
\end{table}

Next, we investigate whether the decomposition method changes the game AI's performance. We allow two game AIs to play the game, one with the evaluation function approximated by simple SVD and the other with the evaluation function approximated by HOSVD. The compression ratio of the evaluation function is controlled by $r$, which is the rank of the approximated matrix. We choose the value of $r$ so that the compression ratio of $f_{\mathrm{SVD}}$ and $f_{\mathrm{HOSVD}}$ are equal. The values of the ranks and the compression ratios are summarized in Table~\ref{table:compression}. For example, in the compression ratio $0.8$ column, the simple SVD has $49$ singular values while the HOSVD has $24$. Since $f_{\mathrm{SVD}}$ can be expressed as $\tilde{Q}\tilde{S}$ and the matrices $\tilde{Q}$ and $\tilde{S}$ contain $r(3^4+3^5)$ elements in total, $f_{\mathrm{SVD}}$ contains $15~876$ elements when $r=64$. Similarly, $f_{\mathrm{HOSVD}}$ contains $3^4r + r^3$ elements, which is $15~768$ when $r=24$. Since the original evaluation function $f_\mathrm{all}$ contains $3^9 = 19~683$ elements, the corresponding compression ratios for the simple SVD and HOSVD are $0.807$ and $0.801$, respectively. While they are slightly different from each other, we identified them both as $0.8$. For the other compression ratios, we chose the number of singular values so that the compression ratios are as similar as possible.

\begin{figure}[htbp]
    \centering
    \includegraphics[width=15cm]{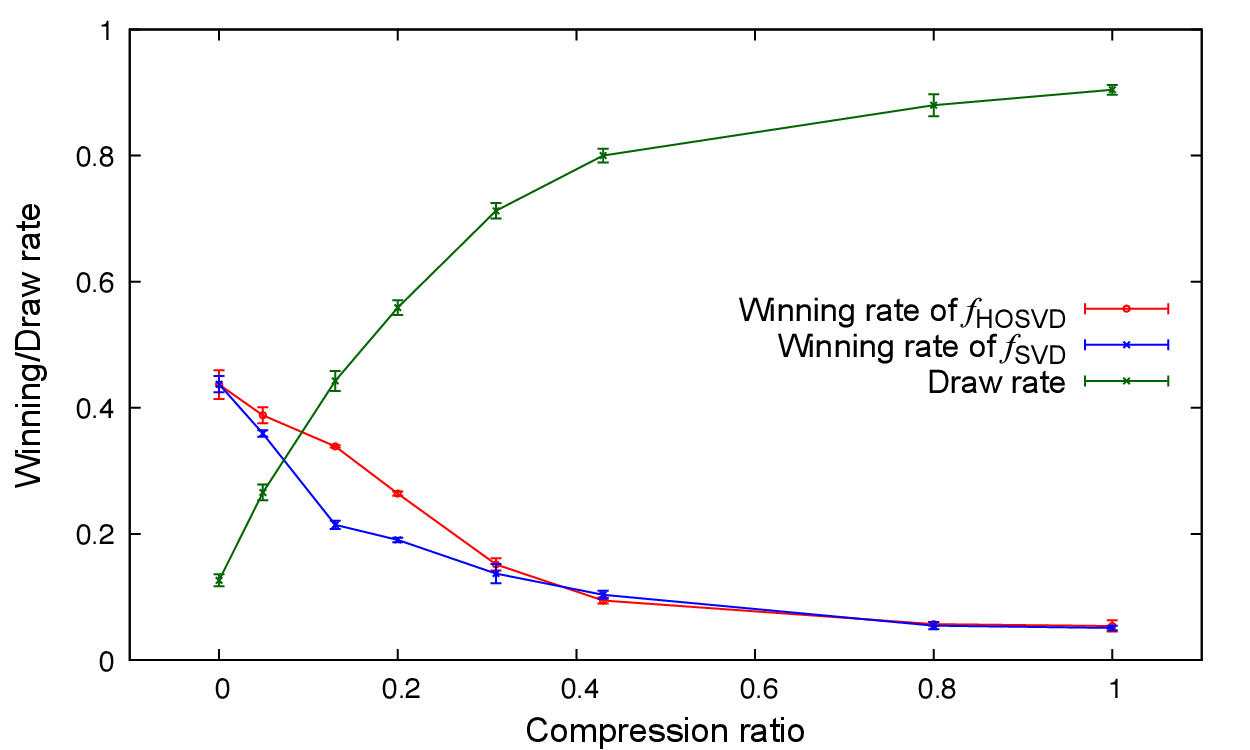}
    \caption{Results of the games between $f_{\mathrm{SVD}}$ and $f_{\mathrm{HOSVD}}$. The red and blue graphs show the winning rate of each evaluation function. The green graph shows the draw rate. When the compression ratio is between 0 and 0.3, the winning rate of HOSVD is significantly larger than that of SVD.}
    \label{fig:res2}
\end{figure}

The winning and draw rates of $f_{\mathrm{SVD}}$ and $f_{\mathrm{HOSVD}}$ are shown in Fig.~\ref{fig:res2}. When the compression ratio is close to $1$, most games are draws, indicating that there is little difference between the two strategies. On the other hand, the winning rate of both strategies becomes $0.5$ when the compression ratio is close to $0$, which means that the two game AIs choose the next moves randomly. When $0 < Cr < 0.3$, the game AI with the evaluation function $f_{\mathrm{HOSVD}}$ exhibits a significantly higher winning rate. This result indicates that the approximation accuracy of HOSVD is greater than that of SVD, which is reflected in the winning rate.

\section{Summary and Discussion}\label{sec:summary}

We performed a low-rank approximation of the evaluation function regarding game board information as a tensor. As a first step to extract features from a game board non-empirically, we studied Tic-Tac-Toe, in which we can construct the perfect evaluation function. We performed low-rank approximation by considering the perfect evaluation function as a ninth-order tensor and investigated the performance of game AIs with approximated evaluation functions. We found that we could reduce the amount of the information of evaluation function by 70\% without significantly degrading the winning rate. As the rank of the approximated evaluation function decreases, the winning rate of the AI with the perfect evaluation function increases. However, the winning rate is not perfectly proportional to the approximation error. This result means that the performance of the approximated evaluation function does not depend only on the singular value distribution. We also investigated the performance of two game AIs with evaluation functions approximated by two different approximation methods: simple SVD and HOSVD. The evaluation function of Tic-Tac-Toe is defined on $3 \times 3$ cells, and the low-rank approximation by simple SVD corresponds to dividing the board into four and five cells, whereas HOSVD divides it into three rows of three cells. Although there was little difference in winning rate when the compression ratio was close to 0 or 1, HOSVD significantly outperformed simple SVD for intermediate values. Since the purpose of the game is to place three marks in a horizontal, vertical, or diagonal row, the decomposition by HOSVD more closely preserves the game's structure than simple SVD. Therefore, it is reasonable that HOSVD has superior performance to simple SVD at the same compression ratio.

Our ultimate goal is to find a non-empirical way to compress the information on the board to pass on to the AI. In this study, we compressed the information of the evaluation function itself by SVD for the first step. We believe that our approach using SVD could be applied to feature learning. However, the specific implementation method is a topic for future study.

\begin{acknowledgment}
    This work was supported by JSPS KAKENHI Grant Number 21K11923.
\end{acknowledgment}

\bibliographystyle{jpsj}
\bibliography{reference}

\end{document}